\documentclass{article}
\usepackage{amsmath}

\usepackage{arxiv}

\usepackage{tabularx}       
\usepackage{booktabs}       
\usepackage{multirow}       
\usepackage{longtable}      
\usepackage{makecell}       
\usepackage{stfloats}       
\usepackage{lipsum}         
\usepackage{booktabs, tabularx}
\usepackage{pifont}
\usepackage{wasysym}   
\usepackage{graphicx}   
\usepackage[table]{xcolor}
\usepackage{colortbl}
\definecolor{yes}{RGB}{198,239,206}
\definecolor{no}{RGB}{255,199,206}
\newcolumntype{L}{>{\raggedright\arraybackslash}X}  

\usepackage[utf8]{inputenc} 
\usepackage[T1]{fontenc}    
\usepackage{hyperref}       
\usepackage{url}            
\usepackage{booktabs}       
\usepackage{amsfonts}       
\usepackage{nicefrac}       
\usepackage{microtype}      
\usepackage{lipsum}

\graphicspath{ {./images/} }

\title{HAWK: A Hierarchical Workflow Framework for Multi-Agent Collaboration}

\author{
  Yuyang Cheng\thanks{These authors contributed equally to this work.} \\
  School of Cyber Science and Engineering\\
  Sichuan University\\
  \texttt{yuyangc125@gmail.com} \\
  \And
  Yumiao Xu\footnotemark[1] \\
  School of Cyber Science and Engineering\\
  Sichuan University\\
  \texttt{1811484216@qq.com} \\
  \And
  Chaojia Yu \\
  School of Computer Science\\
  Sichuan University\\
  \texttt{yuchaojia82@gmail.com} \\
  \And
  Yong Zhao \\
  School of Computer Science\\
  Sichuan University\\
  \texttt{yong.zhao@scupi.cn} \\
}

\begin{document}
\maketitle
\begin{abstract}
Contemporary multi-agent systems encounter persistent challenges in cross-platform interoperability, dynamic task scheduling, and efficient resource sharing. Agents with heterogeneous implementations often lack standardized interfaces; collaboration frameworks remain brittle and hard to extend; scheduling policies are static; and inter‑agent state synchronization is insufficient. We propose \textbf{Hierarchical Agent Workflow (HAWK)}, a modular framework comprising five layers—User, Workflow, Operator, Agent, and Resource—and supported by sixteen standardized interfaces. HAWK delivers an end‑to‑end pipeline covering task parsing, workflow orchestration, intelligent scheduling, resource invocation, and data synchronization. At its core lies an adaptive scheduling and optimization module in the Workflow Layer, which harnesses real‑time feedback and dynamic strategy adjustment to maximize utilization. The Resource Layer provides a unified abstraction over heterogeneous data sources, large models, physical devices, and third‑party services\&tools, simplifying cross‑domain information retrieval. We demonstrate HAWK’s scalability and effectiveness via CreAgentive, a multi‑agent novel‑generation prototype, which achieves marked gains in throughput, lowers invocation complexity, and improves system controllability. We also show how hybrid deployments of large language models integrate seamlessly within HAWK, highlighting its flexibility. Finally, we outline future research avenues—hallucination mitigation, real‑time performance tuning, and enhanced cross‑domain adaptability—and survey prospective applications in healthcare, government, finance, and education.
\end{abstract}


\section{Introduction}
An AI agent is an intelligent entity endowed with autonomous perception, decision‑making, and execution capabilities. It ingests inputs, processes information, and executes appropriate actions within a defined environment to accomplish predetermined tasks. Core attributes of an AI agent include autonomy, interactivity, proactivity, sociability, and the capacity for continual learning and optimization \cite{deng2025ai}.

As AI technology advances toward Artificial General Intelligence (AGI), AI agents have found broad application across domains such as intelligent customer service, autonomous driving, medical diagnostics, smart manufacturing, and financial analysis. The evolution of AI agents can be characterized in four stages:

\begin{itemize}

\item \textbf{Rule-based expert systems:} Early AI agents relied on predefined rules and logical reasoning to perform tasks, such as early medical diagnostic systems.

\item \textbf{Learning‑Driven Intelligent Agents:} The advent of machine learning and deep learning enabled agents to acquire skills autonomously, exemplified by reinforcement‑learning game agents and personalized recommender systems.

\item \textbf{Multi-Agent Collaboration:} Recent research emphasizes coordination and task allocation among multiple agents, as seen in autonomous vehicle fleets and robotic swarms.

\item \textbf{Towards Generalized AI Agents:} Future agents are expected to perform cross‑domain tasks, execute efficient scheduling, and optimize resources in a generalized, adaptive manner.

\end{itemize}

Despite their widespread deployment across multiple industries, AI agent coordination and scheduling technologies remain in their infancy and confront four principal challenges:

\begin{itemize}

\item \textbf{Heterogeneous, multi-vendor systems lacking unified definitions and interfaces:} AI agents developed by different vendors and research institutions often adopt incompatible architectures, communication protocols, and data formats, making direct interaction among agents difficult.The absence of standardized interfaces impedes cross-platform and cross-vendor interoperability, hindering the development and scalability of the AI agent ecosystem.

\item \textbf{Fragmented and rudimentary collaboration frameworks without unified standards:} Most current collaboration frameworks for AI agents are proprietary and domain-specific, designed by individual organizations or enterprises. For example, customer service agents may effectively handle textual dialogues but cannot coordinate with manufacturing agents to complete complex industrial tasks. Due to the lack of a unified collaborative architecture, existing systems exhibit poor scalability, compatibility, and generalizability, falling short of large-scale deployment requirements.

\item \textbf{Lack of Intelligent Scheduling and Optimization Mechanisms:} Current scheduling approaches largely rely on manually defined rules or static allocation strategies, which makes it difficult to adapt to dynamic environments. Task assignment frequently lacks responsiveness to variations in resource constraints, task characteristics, and environmental conditions, and typically depends on manual intervention or simplistic algorithms, thereby limiting overall system adaptability and performance.

\item \textbf{Insufficient intelligent interaction and data state sharing among agents:} Many AI agents function in isolation, lacking mechanisms for cross-agent state synchronization and data sharing. This limitation hinders collaborative task execution and prevents seamless handovers between agents. Furthermore, these agents often possess constrained memory and storage capabilities, which makes it difficult to retain and share historical interaction records. As a result, achieving efficient data exchange, knowledge sharing, and task decomposition in distributed agent networks remains a pressing research challenge.

\end{itemize}

To address the challenges outlined above, this paper proposes a structured multi-agent workflow framework: \textbf{Hierarchical Agent WorKflow (HAWK)}. HAWK is designed to encompass all major aspects of an agent-based system, ranging from client-side workflow specification and submission, through task scheduling and execution, to resource management and provisioning. The framework consists of four functional layers—\emph{User Layer}, \emph{Workflow Layer}, \emph{Operator Layer}, and \emph{Agent Layer}—plus the \emph{Resource Layer}, and defines sixteen key interfaces to ensure both interoperability and flexibility among its components. By decoupling responsibilities across layers, HAWK delivers a modular and independently evolvable architecture that enables developers to focus on business logic without being burdened by underlying heterogeneity or security concerns. Meanwhile, the HAWK framework offers strong compatibility, allowing seamless integration with a variety of communication protocols and retrieval-augmented generation (RAG) frameworks that satisfy its functional requirements. This further enhances the system’s extensibility and intelligence.

Beyond providing a secure and controllable technical solution for multi-agent collaboration, HAWK also establishes a solid foundation for future technological advancements. We demonstrate its applicability and scalability through CreAgentive, a multi-agent novel-generation prototype developed on the HAWK.

\section{Related Work}

\begin{table*}[!htb]
  \centering
  \small
  \caption{Capability Heatmap Across Agent Workflow Systems}
  \label{tab:heatmap_capabilities}
  \begin{tabularx}{\textwidth}{@{}>{\centering\arraybackslash}m{2.66cm} *{9}{>{\centering\arraybackslash}m{1.105cm}}@{}}
    \toprule
    \rowcolor{gray!30}
    \textbf{System} 
    & {\fontsize{7.5}{9}\selectfont \textbf{Planning}}
    & {\fontsize{7.5}{9}\selectfont \textbf{Tool Use}} 
    & {\fontsize{7.5}{9}\selectfont \textbf{Multi-agent}}
    & {\fontsize{7.5}{9}\selectfont \textbf{Memory}} 
    & {\fontsize{7.5}{9}\selectfont \textbf{GUI}}
    & {\fontsize{7.5}{9}\selectfont \textbf{Self-Reflect}} 
    & {\fontsize{7.5}{9}\selectfont \textbf{Custom Tools}} 
    & {\fontsize{7.5}{9}\selectfont \textbf{Cross-Platform}} 
    & {\fontsize{7.5}{9}\selectfont \makecell{\textbf{Distributed}\\\textbf{Exec.}}} \\
    \midrule
    AgentUniverse \cite{agentuniverse2024}
      & \cellcolor{yes}\checkmark 
      & \cellcolor{yes}\checkmark 
      & \cellcolor{yes}\checkmark 
      & \cellcolor{yes}\checkmark 
      & \cellcolor{yes}\checkmark 
      & \cellcolor{no}\Circle 
      & \cellcolor{yes}\checkmark 
      & \cellcolor{yes}\checkmark 
      & \cellcolor{no}\ding{55} \\

    Agentverse \cite{chen2023agentverse}
      & \cellcolor{yes}\checkmark 
      & \cellcolor{yes}\checkmark 
      & \cellcolor{yes}\checkmark 
      & \cellcolor{yes}\checkmark 
      & \cellcolor{no}\RIGHTcircle 
      & \cellcolor{yes}\checkmark 
      & \cellcolor{yes}\checkmark 
      & \cellcolor{yes}\checkmark 
      & \cellcolor{no}\ding{55} \\

    Agno \cite{agno2024}
      & \cellcolor{yes}\checkmark 
      & \cellcolor{yes}\checkmark 
      & \cellcolor{yes}\checkmark 
      & \cellcolor{yes}\checkmark 
      & \cellcolor{no}\ding{55} 
      & \cellcolor{yes}\checkmark 
      & \cellcolor{yes}\checkmark 
      & \cellcolor{yes}\checkmark 
      & \cellcolor{no}\ding{55} \\

    AutoGen \cite{wu2023autogen}
      & \cellcolor{yes}\checkmark 
      & \cellcolor{yes}\checkmark 
      & \cellcolor{yes}\checkmark 
      & \cellcolor{yes}\checkmark 
      & \cellcolor{no}\ding{55} 
      & \cellcolor{yes}\checkmark 
      & \cellcolor{yes}\checkmark 
      & \cellcolor{no}\RIGHTcircle 
      & \cellcolor{yes}\checkmark \\

    CAMEL \cite{li2023camel}
      & \cellcolor{yes}\checkmark 
      & \cellcolor{yes}\checkmark 
      & \cellcolor{yes}\checkmark 
      & \cellcolor{yes}\checkmark 
      & \cellcolor{no}\ding{55} 
      & \cellcolor{yes}\checkmark 
      & \cellcolor{yes}\checkmark 
      & \cellcolor{yes}\checkmark 
      & \cellcolor{no}\ding{55} \\

    ChatDev \cite{qian2023chatdev}
      & \cellcolor{yes}\checkmark 
      & \cellcolor{yes}\checkmark 
      & \cellcolor{yes}\checkmark 
      & \cellcolor{yes}\checkmark 
      & \cellcolor{no}\ding{55} 
      & \cellcolor{no}\RIGHTcircle 
      & \cellcolor{yes}\checkmark 
      & \cellcolor{yes}\checkmark 
      & \cellcolor{no}\ding{55} \\

    Coze \cite{coze2024}
      & \cellcolor{yes}\checkmark 
      & \cellcolor{yes}\checkmark 
      & \cellcolor{yes}\checkmark 
      & \cellcolor{yes}\checkmark 
      & \cellcolor{yes}\checkmark 
      & \cellcolor{no}\Circle 
      & \cellcolor{yes}\checkmark 
      & \cellcolor{yes}\checkmark 
      & \cellcolor{no}\ding{55} \\

    CrewAI \cite{crewai2024}
      & \cellcolor{yes}\checkmark 
      & \cellcolor{yes}\checkmark 
      & \cellcolor{yes}\checkmark 
      & \cellcolor{no}\Circle 
      & \cellcolor{no}\ding{55} 
      & \cellcolor{yes}\checkmark 
      & \cellcolor{yes}\checkmark 
      & \cellcolor{no}\RIGHTcircle 
      & \cellcolor{no}\ding{55} \\

    DeepResearch \cite{openai2025deepresearch}
      & \cellcolor{yes}\checkmark 
      & \cellcolor{yes}\checkmark 
      & \cellcolor{no}\Circle 
      & \cellcolor{yes}\checkmark 
      & \cellcolor{yes}\checkmark 
      & \cellcolor{no}\RIGHTcircle 
      & \cellcolor{yes}\checkmark 
      & \cellcolor{yes}\checkmark 
      & \cellcolor{no}\ding{55} \\

    Dify \cite{langgenius_dify2025}
      & \cellcolor{yes}\checkmark 
      & \cellcolor{yes}\checkmark 
      & \cellcolor{no}\ding{55} 
      & \cellcolor{yes}\checkmark 
      & \cellcolor{no}\ding{55} 
      & \cellcolor{no}\Circle 
      & \cellcolor{yes}\checkmark 
      & \cellcolor{yes}\checkmark 
      & \cellcolor{no}\ding{55} \\

    DSPy \cite{khattab2023dspy}
      & \cellcolor{yes}\checkmark 
      & \cellcolor{yes}\checkmark 
      & \cellcolor{no}\Circle 
      & \cellcolor{yes}\checkmark 
      & \cellcolor{yes}\checkmark 
      & \cellcolor{yes}\checkmark 
      & \cellcolor{yes}\checkmark 
      & \cellcolor{yes}\checkmark 
      & \cellcolor{no}\ding{55} \\

    ERNIE-agent \cite{baidu_ernie_agent2023}
      & \cellcolor{yes}\checkmark 
      & \cellcolor{yes}\checkmark 
      & \cellcolor{yes}\checkmark 
      & \cellcolor{no}\Circle 
      & \cellcolor{yes}\checkmark 
      & \cellcolor{no}\ding{55} 
      & \cellcolor{yes}\checkmark 
      & \cellcolor{yes}\checkmark 
      & \cellcolor{no}\ding{55} \\

    Flowise \cite{flowise2024}
      & \cellcolor{yes}\checkmark 
      & \cellcolor{yes}\checkmark 
      & \cellcolor{no}\ding{55} 
      & \cellcolor{yes}\checkmark 
      & \cellcolor{no}\ding{55} 
      & \cellcolor{no}\ding{55} 
      & \cellcolor{yes}\checkmark 
      & \cellcolor{yes}\checkmark 
      & \cellcolor{no}\Circle \\

    LangGraph \cite{duan2024langgraph_crewai}
      & \cellcolor{yes}\checkmark 
      & \cellcolor{yes}\checkmark 
      & \cellcolor{yes}\checkmark 
      & \cellcolor{yes}\checkmark 
      & \cellcolor{no}\ding{55} 
      & \cellcolor{yes}\checkmark 
      & \cellcolor{yes}\checkmark 
      & \cellcolor{no}\RIGHTcircle 
      & \cellcolor{no}\ding{55} \\

    Magnetic-One \cite{fourney2024magenticone}
      & \cellcolor{yes}\checkmark 
      & \cellcolor{yes}\checkmark 
      & \cellcolor{yes}\checkmark 
      & \cellcolor{yes}\checkmark 
      & \cellcolor{yes}\checkmark 
      & \cellcolor{yes}\checkmark 
      & \cellcolor{yes}\checkmark 
      & \cellcolor{no}\RIGHTcircle 
      & \cellcolor{no}\Circle \\

    Meta-GPT \cite{hong2023metagpt}
      & \cellcolor{yes}\checkmark 
      & \cellcolor{yes}\checkmark 
      & \cellcolor{yes}\checkmark 
      & \cellcolor{no}\RIGHTcircle 
      & \cellcolor{no}\ding{55} 
      & \cellcolor{no}\ding{55} 
      & \cellcolor{yes}\checkmark 
      & \cellcolor{yes}\checkmark 
      & \cellcolor{no}\Circle \\

    n8n \cite{n8n2025}
      & \cellcolor{yes}\checkmark 
      & \cellcolor{yes}\checkmark 
      & \cellcolor{no}\ding{55} 
      & \cellcolor{yes}\checkmark 
      & \cellcolor{yes}\checkmark 
      & \cellcolor{no}\ding{55} 
      & \cellcolor{yes}\checkmark 
      & \cellcolor{yes}\checkmark 
      & \cellcolor{no}\Circle \\

    OmAgent \cite{zhang2024omagent}
      & \cellcolor{yes}\checkmark 
      & \cellcolor{yes}\checkmark 
      & \cellcolor{no}\ding{55} 
      & \cellcolor{yes}\checkmark 
      & \cellcolor{no}\ding{55} 
      & \cellcolor{yes}\checkmark 
      & \cellcolor{yes}\checkmark 
      & \cellcolor{no}\Circle 
      & \cellcolor{no}\Circle \\

    OpenAI Swarm \cite{openai_swarm2024}
      & \cellcolor{yes}\checkmark 
      & \cellcolor{yes}\checkmark 
      & \cellcolor{yes}\checkmark 
      & \cellcolor{no}\ding{55} 
      & \cellcolor{no}\ding{55} 
      & \cellcolor{no}\ding{55} 
      & \cellcolor{yes}\checkmark 
      & \cellcolor{yes}\checkmark 
      & \cellcolor{no}\Circle \\

    Manus \cite{manus2025}
      & \cellcolor{yes}\checkmark 
      & \cellcolor{yes}\checkmark 
      & \cellcolor{yes}\checkmark 
      & \cellcolor{yes}\checkmark 
      & \cellcolor{yes}\checkmark 
      & \cellcolor{no}\ding{55} 
      & \cellcolor{yes}\checkmark 
      & \cellcolor{yes}\checkmark 
      & \cellcolor{no}\Circle \\

    Qwen-agent \cite{qwen_agent2025}
      & \cellcolor{yes}\checkmark 
      & \cellcolor{yes}\checkmark 
      & \cellcolor{yes}\checkmark 
      & \cellcolor{yes}\checkmark 
      & \cellcolor{yes}\checkmark 
      & \cellcolor{yes}\checkmark 
      & \cellcolor{yes}\checkmark 
      & \cellcolor{yes}\checkmark 
      & \cellcolor{no}\Circle \\

    Semantic Kernel \cite{microsoft_semantickernel2024}
      & \cellcolor{yes}\checkmark 
      & \cellcolor{yes}\checkmark 
      & \cellcolor{no}\ding{55} 
      & \cellcolor{yes}\checkmark 
      & \cellcolor{no}\ding{55} 
      & \cellcolor{no}\ding{55} 
      & \cellcolor{yes}\checkmark 
      & \cellcolor{yes}\checkmark 
      & \cellcolor{no}\Circle \\
    \bottomrule
  \end{tabularx}
\end{table*}

Prior to the emergence of AI Agents, the literature predominantly addressed the coordination, scheduling, and optimization of scientific workflows \cite{lu2009collaborative,kashlev2017big,zhao2007swift,ludascher2006scientific}, commercial workflows, big data and intelligent applications \cite{kashlev2017big}, and Web Services \cite{bai2019lpod,bai2023generic}. In addition, unified standards and frameworks for service‑oriented workflow coordination, scheduling, and optimization were established \cite{zhao2014service}, as were algorithms and strategies for resource scheduling and optimization in cloud computing environments \cite{tian2014optimized}. These foundational studies remain pertinent to the AI Agent domain, furnishing both theoretical underpinnings and practical insights.

At present, research on AI Agent coordination and scheduling is advancing rapidly, and promising application results have been demonstrated in domains such as healthcare \cite{chen2025enhancing}, Intelligent control \cite{li2018adaptive}, transport \cite{wu2020multi}, aerospace, agriculture, industrial production \cite{shi2020survey}. Several leading research institutions and enterprises have proposed diverse multi‑agent architectures and solutions; however, most remain in early exploratory stages, with considerable gaps still to be bridged before mature standards and widespread deployment can be achieved.

In terms of compatibility and integration within multi-agent frameworks, existing platforms exhibit various limitations. OpenAI Swarm \cite{openai_swarm2024} currently supports only OpenAI’s own APIs, making it difficult to seamlessly integrate with other LLM services. Qwen‑Agent \cite{qwen_agent2025} has achieved modular workflow orchestration, but it lacks a cross-platform adaptation layer, requiring additional development effort when migrating across ecosystems. Microsoft’s Semantic Kernel \cite{microsoft_semantickernel2024} emphasizes lightweight design and secure integration with enterprise systems; however, its heavy reliance on customized development significantly increases implementation costs. Baidu’s ERNIE‑Agent \cite{baidu_ernie_agent2023} excels in natural language understanding and reasoning but still requires improvement in cross-task scheduling and dynamic optimization.

In general‑purpose multi‑agent collaboration systems, LangGraph \cite{duan2024langgraph_crewai} utilizes a graph‑based workflow management approach that enables flexible modeling of inter‑agent dependencies; however, as task scales grow, the increasing number of node dependencies becomes difficult to maintain. ByteDance’s Coze \cite{coze2024} supports plugin extensions, concurrent operations and an interactive UI, yet suffers from a high failure rate in high‑concurrency scenarios due to immature plugin capabilities and an underdeveloped log‑monitoring system. MetaGPT \cite{hong2023metagpt}, which leverages standard operating procedures (SOPs) and a task‑decomposition mechanism, is well suited to tightly controlled, process‑driven applications but lacks the flexibility to handle unanticipated tasks. Magnetic‑One \cite{fourney2024magenticone}, limited to orchestrating only five preset agents, lacks the scheduling flexibility required for general‑purpose use. Finally, AutoGen \cite{wu2023autogen} is notable for its powerful multi‑agent conversational capabilities, but its cumbersome configuration process is not user‑friendly for non‑technical users.

In the domain of early‐stage exploration and vertical‐scenario applications, Phidata \cite{phidata2025} remains in its nascent stage: although it supports multimodal agents and workflows, its scheduling capabilities are not yet fully developed. Dify \cite{langgenius_dify2025}, aimed at LLM application development, provides a graphical interface and a comprehensive toolset; however, complex requirements still demand manual extensions, thereby increasing development overhead. OmAgent \cite{zhang2024omagent} specializes in multimodal perception and hardware integration, making it particularly suited for real‐time device scenarios; nonetheless, its strong dependence on specific hardware may become a bottleneck under high‐performance demands. Huawei’s MindSpore‑AI \cite{huawei2023mindspore} primarily focuses on training and inference optimization for individual agents, and has yet to establish a comprehensive multi‑agent collaboration mechanism.

\begin{table}[h]
  \centering
  \caption{Interpretation and Scope Clarification}
  \label{tab: Interpretation and Scope Clarification}
  \begin{tabular}{|c|l|}
    \hline
    \(\checkmark\)    & Support                          \\ \hline
    \ding{55}        & Not Support                      \\ \hline
    \(\RIGHTcircle\)  & Partially Support                \\ \hline
    \(\Circle\)      & Unspecified                      \\ \hline
  \end{tabular}
\end{table}

Drawing on the work of Yu et al. \cite{yu2025survey},We have summarized and compared the functionalities of existing multi-agent frameworks in \autoref{tab:heatmap_capabilities}, with explanations of the symbols used provided in \autoref{tab: Interpretation and Scope Clarification}.

\section{HAWK: Hierarchical Agent WorKflow}

\begin{figure}[!htb]  
	\centering  
	\includegraphics[width=\textwidth]{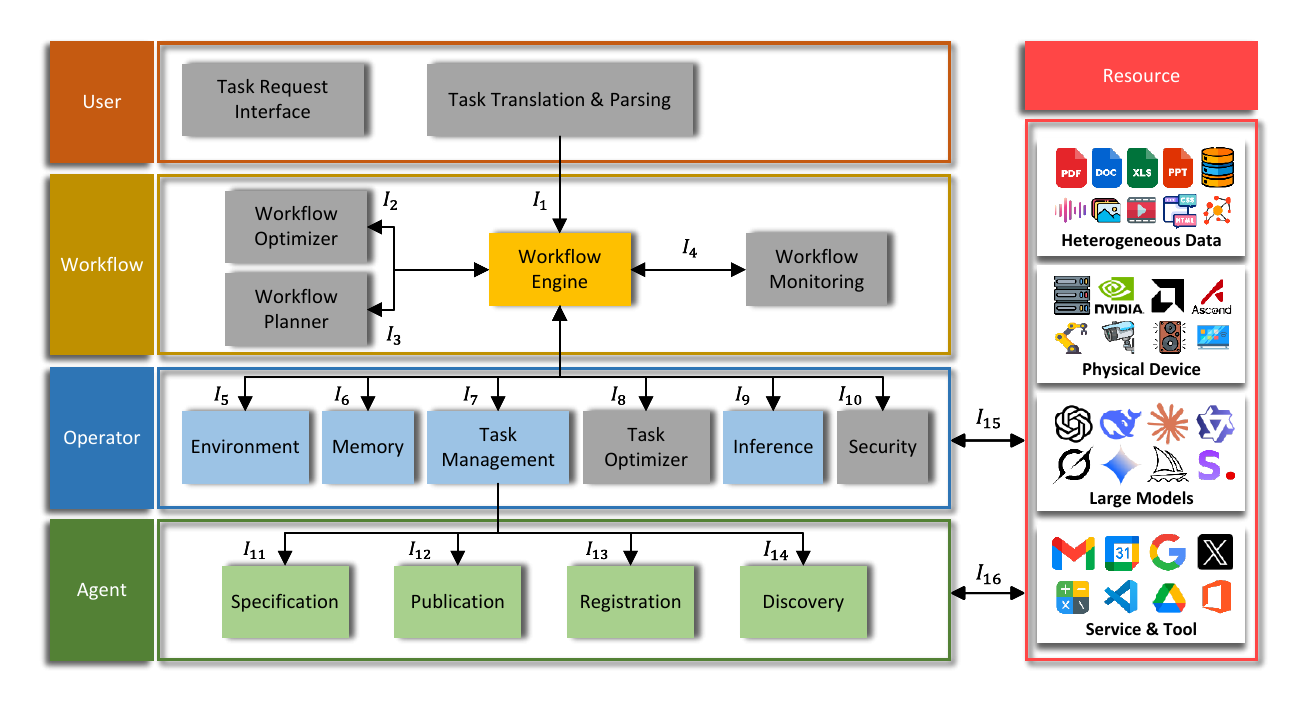}  
	\caption{Framework of Hierarchical Agent WorKflow (HAWK)}
	\label{Fig1: Framework of HAWK}  
\end{figure}

To address the aforementioned challenges, we propose a structured multi-agent workflow framework termed \emph{Hierarchical Agent WorKflow (HAWK)}, which encompasses all critical aspects ranging from client-side workflow specification, agent-based workflow submission and management, task scheduling and execution, to resource management and provisioning. As illustrated in \autoref{Fig1: Framework of HAWK}, HAWK consists of five layers, twenty modules, and sixteen interfaces.

\subsection{Layers}

The first layer is the \emph{User Layer}, which provides users with an interface for submitting task requests, as well as capabilities for task translation and parsing. The processed tasks are then forwarded to the Workflow Layer for subsequent execution. Furthermore, the decoupling of the User Layer from other layers offers flexibility in customizing the system’s user interface and facilitates the reuse of other system components across diverse scientific domains.

The second layer is the \emph{Workflow Layer}, which is responsible for transferring parsed and preprocessed user tasks from the upper layer to the \emph{Workflow Engine}, and for orchestrating the planning, execution, monitoring, and optimization of agent workflows. The \emph{Wrkflow Engine} selects appropriate workflow models based on task requirements and exposes interfaces to the underlying Operator Layer to direct the execution of decomposed tasks. The decoupling of the Workflow Layer offers two main advantages: first, it isolates workflow model selection from task model selection, ensuring that modifications to the workflow structure do not affect task structures; second, it separates workflow scheduling from task execution, thereby enhancing the overall performance and scalability of the management system.

The third layer is the \emph{Operator Layer}, which is mainly responsible for the scheduling and execution of specific tasks within the agent workflow. This layer comprises six core modules: \emph{Environment}, which provides contextual configurations for task operations; \emph{Memory}, which supports state maintenance and historical information management during task execution; \emph{Task Management}, which handles task distribution and monitoring; \emph{Task Optimizer}, which dynamically adjusts execution strategies based on policies and available resources; \emph{Reasoning} module performs complex reasoning during task execution to enhance task accuracy; \emph{Security}, which ensures privilege control and data protection throughout the process. The Operator Layer serves as the central hub that bridges scheduling logic and execution, offering stable, efficient, and controllable support for agent operation.

The fourth layer, referred to as the \emph{Agent Layer}, is responsible for managing four primary tasks related to agents, as issued by the Task Management module of the Operator Layer: \emph{Specification}, \emph{Publication}, \emph{Registration}, and \emph{Discovery}. Decoupling the Agent Layer from the Operator Layer enables the latter to maintain cohesion and controllability over its overall scheduling strategy, while allowing agents to be deployed flexibly across diverse devices or containers, thereby adapting to heterogeneous environments. This clear separation of responsibilities facilitates the construction of an autonomous and collaborative multi-agent architecture, significantly enhancing system robustness, fault tolerance, and execution efficiency.

The \emph{Resource Layer}, positioned on the right side of the framework, serves as the fundamental support layer for the multi-agent workflow system, responsible for supplying the Operator Layer and Agent Layer with various resources. This layer primarily encompasses four categories: \emph{Heterogeneous Data}, including both structured and unstructured data; \emph{Large Models}, comprising diverse pre-trained or fine-tuned language and image models, etc.; \emph{Physical Devices}, such as cameras, robotic arms, and other hardware capable of direct environmental interaction; \emph{Tools \& Services}, which cover third-party APIs and algorithm libraries. Moreover, through a unified resource access and abstraction mechanism, the Resource Layer delivers standardized interfaces to both the Operator and the Agent Layers, effectively lowering the invocation threshold and enhancing system compatibility and scalability.

\subsection{Interfaces}

In the reference architecture, 16 interfaces are explicitly defined to specify how each module interacts with others. Interoperability among modules is ensured through the standardization of these interfaces. Such standardized interfaces enable end-to-end coordination—from user requests and resource invocation to the autonomous collaboration of agents.

Interface \emph{I\textsubscript{1}} defines the communication protocol between the \emph{Task Translation\&Parsing} and the \emph{Workflow Engine}, enabling parsed and normalized tasks to be forwarded for selecting or constructing an appropriate agent workflow to address the given task.

Interface \emph{I\textsubscript{2}} defines the communication protocol between the \emph{Workflow Optimizer} and the \emph{Workflow Engine}, enabling the optimizer to receive real-time feedback on workflow execution and to dynamically adjust workflow structures and task scheduling strategies. This interaction enhances overall system throughput and execution efficiency.

Interface \emph{I\textsubscript{3}} defines the communication protocol between the \emph{Workflow Engine} and the \emph{Workflow Planner}, through which the engine delegates planning-related responsibilities—such as task decomposition, temporal arrangement, and dependency resolution—to the planner. By offloading these functions, the engine simplifies its runtime execution logic while benefiting from more structured and adaptive workflow plans.

Interface \emph{I\textsubscript{4}} defines the communication protocol between the \emph{Workflow Engine} and the \emph{Workflow Monitoring}, enabling real-time reporting of execution states and key performance metrics from individual workflow nodes. This data serves as a critical foundation for subsequent workflow analysis and optimization.

Interfaces \emph{I\textsubscript{5}}–\textit{I\textsubscript{10}} define the communication protocols between the Workflow Engine and the six core modules of the Operator Layer. These protocols enable the Workflow Engine to dispatch workflow tasks to the corresponding execution modules and to collect their results for centralized coordination and holistic workflow management.

Interfaces \emph{I\textsubscript{11}}–\textit{I\textsubscript{14}} define the communication protocols between the Task Management module and four fundamental agent operations: Specification, Publication, Registration, and Discovery. These protocols enable the Task Management module to interact with agent instances in a standardized manner, thereby facilitating resource allocation, agent deployment, and coordinated task execution.

Interface \emph{I\textsubscript{15}} defines the communication protocol between the Operator Layer and the Resource Layer. This protocol enables structured and uniform access to underlying resources during task execution and supports the logging and storage of data generated throughout the execution process.

Interface \emph{I\textsubscript{16}} defines the communication protocol between the Agent Layer and the Resource Layer. This protocol enables agents to dynamically access various types of resources—such as selecting task-appropriate large language models or external tools—in order to instantiate and publish task-specific agent instances for effective execution.

\subsection{Discussion}

The motivation behind this work arises from the limitations and challenges of existing multi-agent workflow frameworks and products, particularly concerning interoperability, scheduling optimization, and usability. To address these issues, we propose the design of an efficient and scalable multi-agent workflow framework composed of five layers:

\begin{itemize}

\item \textbf{User Layer:} Receive user task requests via APIs or visual interfaces and translate natural language or graphical input into structured descriptions;

\item  \textbf{Workflow Layer:} Responsible for process engine, monitoring, planning and global optimization for high level task disassembly, dependency management and performance improvement;

\item \textbf{Operator Layer:} It consists of six modules: Environment, Memory, Task Management, Task Optimization, Reasoning and Security, which provide fine-grained scheduling, context maintenance and runtime dynamic optimization of Agent tasks;

\item \textbf{Agent Layer:} Implement Specification, Publication, Registration, Discovery four Agent governance mechanisms to support the dynamic publication, registration, discovery and collaboration of intelligent bodies;

\item \textbf{Resource Layer:} Aggregate heterogeneous data, large models, physical devices and tool services to provide a unified resource abstraction and access interface.

\end{itemize}

By decoupling responsibilities across distinct layers, HAWK achieves modular abstraction and and supports independent evolution, enabling developers to concentrate on core business logic without being hindered by underlying heterogeneity or security challenges. Its sixteen standardized interfaces orchestrate both inter-layer data and command flows and inter-layer governance, offering high flexibility and extensibility. These interfaces can be further extended to accommodate diverse application scenarios, such as resource access, task scheduling, and collaboration strategies, resulting in a stable, controllable multi-layer workflow solution that can evolve with future technological advances.

In the current multi‐agent ecosystem, several widely adopted industry standards have emerged, including the Message-Centric Protocol (MCP) \cite{anthropic2024mcp} for loose coupling, the Agent-Network Protocol (ANP) \cite{anp2025} for lifecycle and topology management, and Agent-to-Agent communication (A2A) \cite{google2025a2a} for peer-to-peer negotiation. HAWK is positioned above these protocols, offering a higher-level, unified abstraction. Adopting a design philosophy of “top-layer coverage and abstract standardization,” HAWK not only fully encompasses the core capabilities of MCP, ANP, and A2A, but also exposes extensive extension points to accommodate future protocol iterations and feature evolution. Notably, industry vendors have already begun optimizing their systems  in alignment with our layered blueprint, thereby validating both the theoretical guidance and practical feasibility of the HAWK framework. HAWK can also be seamlessly integrated with various retrieval-augmented generation (RAG) frameworks currently available, enabling efficient retrieval and reasoning over distributed, heterogeneous knowledge sources and further enhancing the system's intelligence and responsiveness. 

Furthermore, HAWK can be seamlessly integrated within both the medical domain (such as medical IoT, electronic health records, clinical workflow coordination) and intelligent robotics (such as human-robot interaction, collaborative robot swarms). In the \emph{Resource Layer}, heterogeneous devices like medical imaging systems, vital sign monitors, and surgical manipulators are unified under a common abstraction. The \emph{Workflow Layer} and the \emph{Operator Layers} provide real‐time monitoring, planning and dynamic optimization, ensuring end-to-end quality of service in critical scenarios such as surgical procedures and emergency dispatch. Within the \emph{Agent Layer}, our governance mechanisms enable robot teams to execute standardized protocols for tasks like surgical assistance, environmental inspection, and rehabilitation support, while collaborating seamlessly with clinical staff. Consequently, HAWK transcends a purely academic construct and serves as a robust, multi-layered framework adaptable to diverse industrial applications, with “medical + robotics” representing one of its many high-impact use cases.

\section{IMPLEMENTATION: CreAgentive}

\begin{figure*}[!htb]  
	\centering  
	\includegraphics[width=\textwidth]{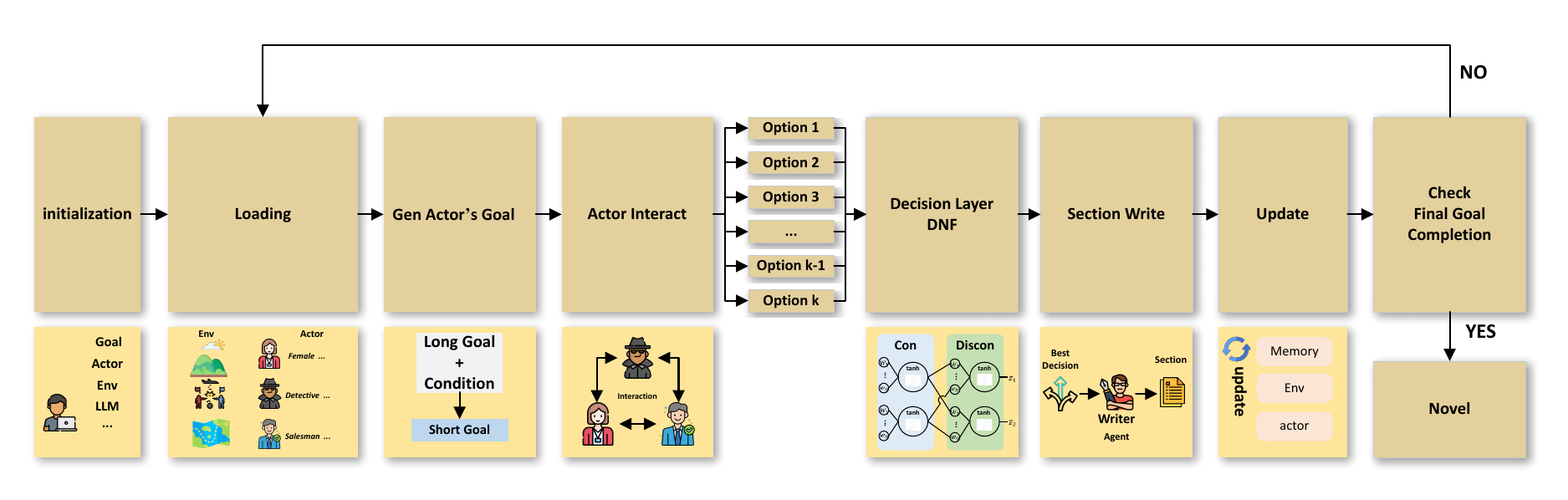}  
	\caption{The Workflow of the CreAgentive}
	\label{Fig2:Demo Workflow}  
\end{figure*}

In this section, we demonstrate the applicability of HAWK by implementing a multi-agent, AI-driven text generation prototype named \textbf{CreAgentive}. Drawing inspiration from established systems such as Drama Engine \cite{pichlmair2024dramaengineframeworknarrative} and Multiagent-poetry \cite{zhang2024llmbasedmultiagentpoetrygeneration}, CreAgentive showcases how HAWK can effectively orchestrate a collection of specialized agents to collaboratively compose a novel.

The workflow of CreAgentive is structured into several stages, each dedicated to a specific aspect of the novel-writing process. The system operates iteratively in a loop until the novel reaches its predefined conclusion.

\subsection{Workflow}

The whole system iteratively generates chapters within a multi-agent environment by the coordinated interplay of long‑term and short‑term goals, continuing until the predefined narrative conclusion is reached. \autoref{Fig2:Demo Workflow} presents the workflow schematic, showing a closed loop from initialization through to ending determination.

First, the system performs the necessary resource preparation and initializes its state variables. Specifically, it locates and loads the environment description files, character profiles and memory archives, and the master story outline. At this point, the system is ready to enter the main iterative loop.

CreAgentive employs a two‑tiered goal hierarchy:

\begin{itemize}

    \item \textbf{Long‑term goals} are derived directly from the predefined outline and serve to guide the overall narrative trajectory and to determine when the story has reached its conclusion.

    \item \textbf{Short‑term goals} are generated immediately prior to each chapter.  Here, each agent reasons over the current environment state and its own historical memory in light of the long‑term goals; using a chain‑of‑thought (CoT) approach, it produces a chapter‑specific objective and refines this into a concrete, step‑by‑step plan. This closed loop of goal generation and planning ensures that each chapter advances the story in a coherent, goal‑driven manner.

\end{itemize}

The core of the workflow is the main loop, which repeats until the ending condition is satisfied. Under the predicate “ending not yet complete,” the system proceeds as follows:

\begin{enumerate}
    \item \textbf{Environment and Character Loading.}  The latest versioned environment file and current character states are loaded into memory.

    \item \textbf{Short‑term Goal and Plan Generation:}  Each agent generates a short-term objective by reasoning over the current environment and its internal memory. Based on this goal, it constructs a detailed action plan. The agents then execute their plans within the shared environment, resulting in multiple candidate story trajectories stemming from a common initial state.

    \item \textbf{Decision‑Level Selection:}  The pool of candidate trajectories is passed to the Decision Agent, which, using a Teller‑inspired cognitive–decision architecture \cite{liu-etal-2024-teller}, evaluates and selects the trajectory that best satisfies our narrative criteria.

    \item \textbf{Automated Chapter Writing:} The chosen trajectory is handed to the Writer Agent. This specialist agent constructs a tailored prompt and invokes the LLM to render the full chapter.

    \item \textbf{Versioned State Update:} Upon completing a chapter, the system updates and assigns version tags to both the environment state and character memories, thereby maintaining a comprehensive audit trail of the story’s progression.

    \item \textbf{Ending Check:} The system leverages the LLM to evaluate the current narrative state against the predefined outline. If the ending condition is satisfied, the iteration loop terminates; otherwise, the process continues with the generation of the next chapter.

\end{enumerate}

This structured and goal-driven workflow ensures that each chapter contributes meaningfully to both the micro‑level continuity and the macro‑level arc of the novel, while maintaining full traceability and controllability throughout the creative process.

\subsection{Key Components}

This subsection introduces the four core functional components of our novel-writing workflow: the Environment Agent, the Decision Agent, the Writer Agent, and the Ending Determination Agent.

\begin{figure}[!h]  
	\centering  
	\includegraphics[width=\linewidth]{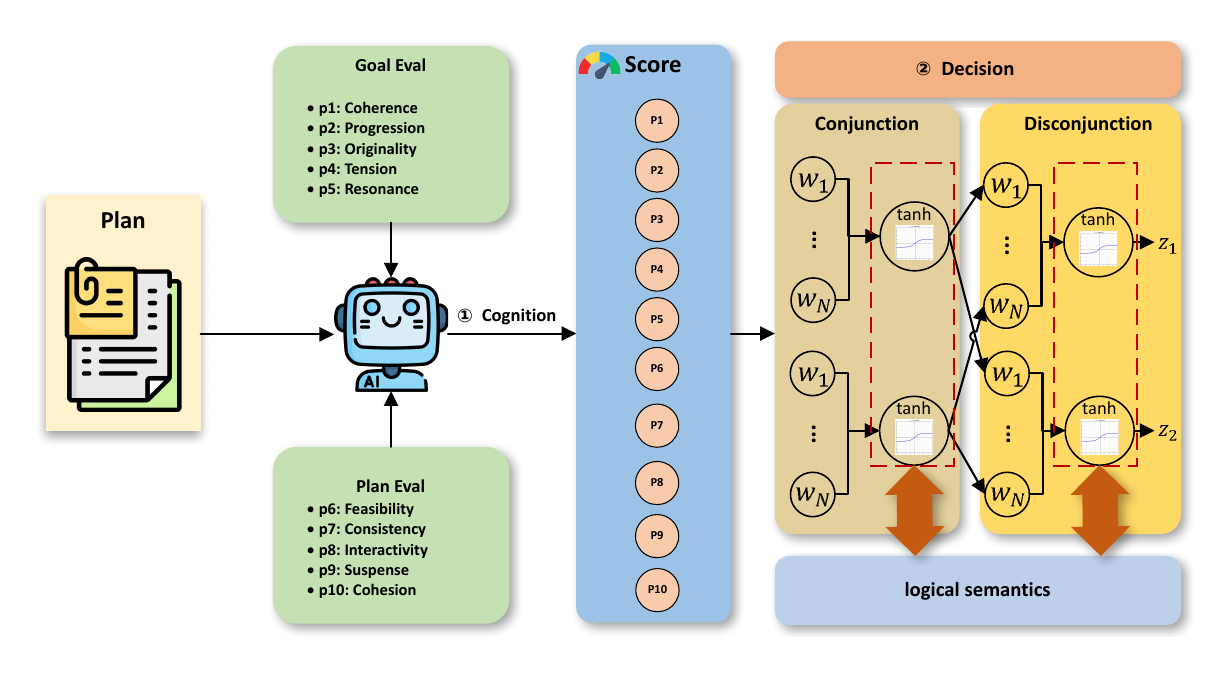}  
	\caption{The Framework of Decision Agent}
	\label{Decision}  
\end{figure}

\begin{itemize}
    \item \textbf{Environment Agent:} Responsible for maintaining the story’s evolving world state, this agent handles the storage, retrieval, and versioned updating of all environment-related data. At the beginning of each iteration, it loads the corresponding version-tagged environment file. Following the execution of an agent-generated action plan, it produces an updated environment state reflecting narrative changes. By preserving each chapter's context under a distinct version identifier, the system enables complete traceability and parallel review of the story’s development over time.

    \item \textbf{Decision Agent:} Building on the Teller dual-system cognitive architecture \cite{liu-etal-2024-teller}, this agent integrates a differentiable Disjunctive Normal Form (DNF) reasoning layer to evaluate and rank alternative candidate storylines. For each predicate $P_i$, the truth value of its $k$-th logic atom is represented as $\mu_{i,k} \in [-1,1]$, computed from either LLM-derived logits or sampled outputs (see Eqs.~\eqref{eq:truth_open} and \eqref{eq:truth_closed}).

The DNF layer constructs $C$ conjunctive clauses:
\[
\mathrm{conj}_c = \bigwedge_{p_{i,k} \in A_c} p_{i,k},
\quad A_c \subseteq \{p_{1,1}, \dots, p_{N,M_N}\},
\]
which are aggregated into per-label disjunctions:
\[
s_y = \bigvee_{c \in C_y} \mathrm{conj}_c,
\quad C_y \subseteq \{1, \dots, C\},
\]
producing confidence scores for each label $y$, which are normalized via softmax:
\[
z_y = \mathrm{softmax}(s_y).
\]
The model is trained end-to-end using the standard cross-entropy loss:
\[
\mathcal{L} = -\sum_{y \in Y} \mathbf{1}(y = y^*) \log z_y.
\]

Specifically, for an \emph{open-logit} LLM, truth values are computed as:
\begin{equation}\label{eq:truth_open}
\mu = 2 \cdot \frac{e^{v_{\mathrm{Yes}}}}{e^{v_{\mathrm{Yes}}} + e^{v_{\mathrm{No}}}} - 1,
\end{equation}
while for a \emph{closed-logit} LLM using $m$ sampled outputs:
\begin{equation}\label{eq:truth_closed}
\mu = 2 \cdot \frac{m_{\mathrm{Yes}}}{m_{\mathrm{Yes}} + m_{\mathrm{No}}} - 1.
\end{equation}

\autoref{Decision} illustrates the overall decision-making pipeline: logical truth values $\mu_{i,k}$ are evaluated through learned conjunctive and disjunctive structures, enabling interpretable, controllable selection of the optimal plot trajectory.

    \item \textbf{Writer Agent:} The Writer Agent is tasked with transforming abstract plans into coherent narrative text. It takes as input the optimal trajectory selected by the Decision Module along with the current state of the environment, constructs a task-specific prompt, and invokes the LLM to generate the complete chapter. This agent serves as the crucial link between decision-level planning and language-level realization.

    \item \textbf{Ending Determination Agent:} At the conclusion of each iteration, this module compares the current environment and character states with the predefined story outline. Using the evaluative capabilities of the LLM, it determines whether the narrative has reached its intended ending. If the ending condition is satisfied, the module signals the termination of the workflow loop; otherwise, the process proceeds to the next iteration.
    
\end{itemize}

\subsection{Discussion}

In the foregoing, we have presented a proof‑of‑concept implementation of our HAWK framework for agent‑based novel creation. \autoref{Fig1: Framework of HAWK} illustrates the implementation status of each module: components shaded in gray remain to be developed. Thanks to HAWK’s hierarchical, decoupled architecture, our team was able to focus primarily on realizing the workflow itself—without being encumbered by the low‑level design and implementation details of a multi‑agent system—thereby greatly accelerating our development process.

The successful execution of our novel‑writing workflow provides preliminary validation of HAWK’s overall correctness. Nevertheless, as with other multi‑agent systems (MAS) reported in the literature \cite{cemri2025multiagentllmsystemsfail}, we encountered two principal challenges:

\begin{enumerate}

    \item \textbf{LLM Hallucinations.} Periodic generation errors or “hallucinations” from the language model can unexpectedly interrupt the workflow.

    \item \textbf{Rule‑Violation in Generated Content.} Although the model generally adheres to the story outline, it sometimes produces narrative elements that conflict with real‑world logic or established story constraints.

\end{enumerate}

We plan to address both issues by implementing the Workflow Monitoring module within HAWK’s Workflow layer. Once all HAWK modules are fully implemented, we anticipate robust, multi‑agent workflows that mitigate hallucinations and enforce narrative consistency.

During development, we also tested three leading LLMs, Deepseek‑V3, Qwen/QwQ‑32B, and GLM‑4‑9B, on identical story‑generation tasks. Our findings indicate:

\begin{itemize}

    \item \textbf{Deepseek‑V3} delivers the highest narrative quality with the fewest hallucinations, reliably producing the greatest number of chapters.
    
    \item \textbf{Qwen/QwQ‑32B} and \textbf{GLM‑4‑9B} outperform Deepseek‑V3 in chain‑of‑thought reasoning for long‑ and short‑term goal generation and in plan formulation, exhibiting comparable performance to each other.
    
\end{itemize}

\begin{table}[h]
\centering
\small
\caption{Comparison of LLM Performance in Multi-Chapter Continuity}
\label{tab:llm-continuity-comparison}
\begin{tabular}{@{}lccc@{}}
\toprule
\textbf{Model} & \textbf{Avg.Chapters}& \textbf{Continuity} & \textbf{Weaknesses}  \\
\midrule
Deepseek-V3   & 20 & Strong     & Over-imaginative\\
Qwen-32B      & 10 & Moderate   & Inconsistent formatting\\
GLM-4-9B      & 6  & Weak       & Less fluent  \\
\bottomrule
\end{tabular}
\end{table}

\autoref{tab:llm-continuity-comparison} summarizes the comparative performance of the tested models in terms of multi-chapter continuity. These results suggest that, in future iterations, a hybrid approach—selecting the optimal model for each specific module—will yield the best overall performance. Beyond multi-chapter continuity, we conducted an internal evaluation of the HAWK workflow along several dimensions:

\begin{itemize}
    \item \textbf{Execution Efficiency:} On average, the complete novel-writing process (including goal-setting, planning, writing, and reviewing) for a 10-chapter story took approximately 80 minutes, with Qwen/QwQ-32B contributing to the fastest generation time.

    \item \textbf{Module Stability:} Across 50 test runs, over 92\% of modules executed without interruption. Most failures originated from hallucinated tool calls or malformed outputs in the Planning and Writing phases.

    \item \textbf{Resource Utilization:} Our workflow exhibited scalable behavior under parallel execution, supporting up to 5 concurrent storyline generations with no observed degradation in quality or timing.
\end{itemize}

These results demonstrate that HAWK not only guarantees strong narrative coherence but also offers practical robustness and fault recovery capabilities. Although our current implementation focused on novel writing, CreAgentive's modular and flexible design makes it readily adaptable to other creative tasks such as screenwriting, interactive storytelling, and game narrative generation, extending its utility across a wide range of content creation scenarios. Furthermore, leveraging the HAWK framework, CreAgentive can be enhanced by constructing an extensive textual knowledge base and integrating retrieval-augmented generation (RAG) techniques. This approach will enable more diverse and innovative content creation while improving controllability and variety. Moreover, our design agent can be extended to various information-retrieval tasks to help retrieve optimal results.

\section{Conclusion}

In this paper, we introduced \textbf{HAWK}, a structured multi-agent workflow framework designed to address key challenges in agent coordination, task scheduling, and resource orchestration. HAWK is composed of five layers—\emph{User}, \emph{Workflow}, \emph{Operator}, \emph{Agent}, and \emph{Resource}—and is underpinned by sixteen standardized interfaces that ensure modularity, extensibility, and interoperability. Its key contributions include:

\begin{itemize}
    \item \textbf{Layered Architecture:} Clear separation of concerns across layers reduces complexity and accelerates development.

    \item \textbf{Standardized Interfaces:} Sixteen Standardized Interfaces ensure seamless interoperability across platforms and vendors.

    \item \textbf{Adaptive Scheduling:} Intelligent Task Management, Reasoning, Optimization, and Security modules enable real-time, context-aware task allocation.

    \item \textbf{Unified Resource Abstraction:} A common access layer for heterogeneous data, models, and services simplifies integration and scales effortlessly.
    
\end{itemize}

To evaluate the applicability and scalability of HAWK, we developed \textbf{CreAgentive}, a multi-agent novel-generation system built on the HAWK architecture. Experimental results demonstrate that HAWK significantly improves task throughput, simplifies resource invocation, and enhances overall system controllability. Comparative evaluations across various large language models further validate the effectiveness of hybrid-model deployments in real-world scenarios. Despite these promising results, the current prototype implementation via CreAgentive still faces several limitations: 1) occasional LLM hallucinations that may disrupt workflow consistency; 2) performance bottlenecks under high-concurrency conditions; 3) insufficient domain-level adaptation, which limits robustness and fault tolerance in complex domains such as healthcare and manufacturing.

Looking ahead, we will deploy and evaluate HAWK in several key domains:

\begin{itemize}

    \item \textbf{Healthcare:} Multi-agent diagnostic systems combining LLMs with medical knowledge for accurate recommendations and efficient resource use.

    \item \textbf{Government Services:} Cross-agency coordination for smart cities, public safety, and policy optimization.

    \item \textbf{Economics and Finance:} High-throughput analysis and forecasting to support informed decision-making.

    \item \textbf{Education:} Personalized learning platforms powered by multimodal content and adaptive feedback.
  
\end{itemize}

Underpinning these deployments is HAWK’s architectural foundation, designed to orchestrate millions of heterogeneous agents across distributed environments. Its high-throughput communication infrastructure and adaptive resource management capabilities facilitate seamless collaboration among diverse AI entities—from industrial controllers and scientific simulators to trading bots and civic service agents. Leveraging a multi-modal collaboration protocol, HAWK enables cross-domain synergy, allowing, for example, medical agents to inform public health policy or educational AIs to adapt to labor market trends. Moreover, HAWK can be extended to a wide range of information retrieval tasks by incorporating retrieval-augmented generation (RAG) mechanisms, enabling efficient querying and knowledge fusion across distributed, heterogeneous data sources. These capabilities collectively enhance the system’s intelligence, adaptability, and responsiveness.

These large-scale implementations will further advance the maturity of HAWK and empirically validate its value within real-world ecosystems. By enabling intelligent, cross-sector integration at scale, HAWK lays the groundwork for a unified, collaborative AI infrastructure spanning industry, science, government, and education.


\section*{Acknowledgments}
This work was funded by the National Natural Science Foundation of
 China (NSFC) under Grant [No.62177007].

\bibliographystyle{unsrt}  
\bibliography{references}

\end{document}